\pdfoutput=1

\documentclass[journal]{IEEEtran}

\usepackage[pdftex]{graphicx}
\graphicspath{{./Figures/}}
\usepackage{epsfig}
\usepackage{graphicx}
\usepackage{algorithmic}
\usepackage{enumitem}
\usepackage{array}
\usepackage{dblfloatfix}

\usepackage{amsmath, bbm, amsthm} 
\usepackage{amssymb}  
\usepackage{arydshln}
\usepackage{todonotes}
\usepackage{url}
\usepackage{tabulary}

\usepackage{amsfonts}
\usepackage{todonotes}
\usepackage[caption=false, font=footnotesize]{subfig}

 \usepackage[utf8]{inputenc} 
\usepackage[T1]{fontenc}    
\usepackage{hyperref}       
\usepackage{url}            
\usepackage{booktabs}       
\usepackage{amsfonts,amssymb,amsmath,amsthm}
\usepackage{nicefrac}       
\usepackage{microtype}      
\usepackage{graphicx}
\usepackage{epsfig}
\usepackage{color}

\usepackage{amsfonts}
\usepackage{float}
\usepackage{url}
\usepackage{epstopdf}
\usepackage{cuted}
\usepackage{tikz}
\usepackage{tabularx}
\usepackage{tabulary}
\usepackage{lipsum}
\usepackage{enumitem}
\usepackage{multirow}
\usepackage{empheq}
\usepackage[most]{tcolorbox}
\usepackage[export]{adjustbox}

\usepackage{tikz}
\usepackage{tikz-qtree}
\usetikzlibrary{trees} 

\newcommand{\R}{\mathbb{R}}

\newcommand{\Ebb}{\mathbb{E}}
\newcommand{\N}{\mathcal{N}}
\newcommand{\E}{\mathcal{E}}
\newcommand{\G}{\mathcal{G}}
\newcommand{\Sc}{\mathcal{S}}

\newcommand{\C}{\mathcal{C}}

\newtheorem{lemma}{Lemma}

\begin{document}
%
\title{Towards Distributed Energy Services: Decentralizing Optimal Power Flow with Machine Learning}
%
%
%

\author{Roel~Dobbe,~Oscar~Sondermeijer,~David~Fridovich-Keil,~Daniel~Arnold, Duncan~Callaway~and~Claire~Tomlin
\thanks{R. Dobbe is with the AI Now Institute at New York University, New York, USA, email: roel@ainowinstitute.org. D Fridovich-Keil and C. Tomlin are with the Department
of Electrical Engineering \& Computer Sciences, UC Berkeley. D. Arnold is with the Grid Integration Group at Lawrence Berkeley National Laboratory. D. Callaway is with the Energy \& Resources Group at UC Berkeley.}
\thanks{Manuscript received June 2018.}}

%
%

\markboth{Submission to Transactions on Smart Grid}%
{Shell \MakeLowercase{\textit{et al.}}: Bare Demo of IEEEtran.cls for IEEE Journals}
%



\maketitle

\begin{abstract}
The implementation of optimal power flow (OPF) methods to perform voltage and power flow regulation in electric networks is generally believed to require extensive communication. 
We consider distribution systems with multiple controllable Distributed Energy Resources (DERs) and present a data-driven approach to learn control policies for each DER to reconstruct and mimic the solution to a centralized OPF problem from solely locally available information. 
Collectively, all local controllers closely match the centralized OPF solution, providing near-optimal performance and satisfaction of system constraints.
A rate distortion framework enables the analysis of how well the resulting fully decentralized control policies are able to reconstruct the OPF solution.
The methodology provides a natural extension to decide what nodes a DER should communicate with to improve the reconstruction of its individual policy.
The method is applied on both single- and three-phase test feeder networks using data from real loads and distributed generators, focusing on DERs that do not exhibit intertemporal dependencies.
It provides a framework for Distribution System Operators to efficiently plan and operate the contributions of DERs to achieve Distributed Energy Services in distribution networks.
\end{abstract}


%
\IEEEpeerreviewmaketitle

%

\section{Introduction}
\label{sec:introduction}

%
%
Many Distribution System Operators (DSOs) are testing ways to prevent expensive network updates by exploiting Distributed Energy Resources (DERs) and their sensing and actuation capabilities to provide \emph{Distributed Energy Services} (DES) to enable \emph{active distribution networks}. 
In DES, DERs are embraced (1) to compensate for the negative effects of distributed generation (DG) and electric vehicle (EV) charging to allow for higher levels of penetration, and (2) to distribute and diversify capital investments on the grid \emph{both in space and time}, to align closer with actual changes that occur organically. The latter benefit can dramatically decrease capital costs, as larger updates tend to be more conservative due to longer and more uncertain planning horizons~\cite{lacey_interchange_2018}.
%
%
%


Earlier efforts to address voltage and power flow problems related to higher penetrations of DERs aimed at developing \emph{decentralized control based on local information}~\cite{carvalho_distributed_2008,keane_enhanced_2011,smith_smart_2011,turitsyn_options_2011}.
These use heuristics to adjust reactive power output at each inverter based on the local voltage, and have shown promise in their ability to reduce voltage variability, but suffer from extensive tuning, which is impractical for larger networks with many inverters. In addition, these methods yield suboptimal control signals and generally do not guarantee the satisfaction of critical system constraints~\cite{cavraro_value_2016}.
A step further, we find methods that consider a control theoretic formulation emulating proportional~\cite{li_real-time_2014} or integral control~\cite{zhang_local_2013}.
Some of these exploit sensitivities between controllable variables and relevant quantities in the network such as critical voltage or branch flow levels~\cite{rogers_authenticated_2010,robbins_two-stage_2013,calderaro_optimal_2014}.
Lastly, \cite{zhang_local_2013} proposes a simple localized optimization framework aiming to track a local voltage reference, and provides necessary conditions that show that these methods do not converge and may not be stable for longer and more heavily loaded networks.

The emerging need for active distribution grids also motivates the use of Optimal Power Flow (OPF) to enable \emph{optimization \& control based on system-wide information}~\cite{farivar_optimal_2012}. 
OPF refers to solving an optimization problem that minimizes some economic or operational objective subject to power flow and other constraints, such as capacity or thermal limits.
Traditionally, OPF is used \emph{offline} as a design tool for network upgrades to size and place equipment, as proposed for capacitor planning in~\cite{baran_optimal_1989}, or as a planning tool to schedule the dispatch of generators and control equipment.
With recent theoretical advances in optimization overcoming a lack of convergence guarantees~\cite{bai_semidefinite_2008,lavaei_zero_2012}, the implementation of OPF in an \emph{online} setting has been a popular research area. 
The inherent lack of communication infrastructure in distribution networks subsequently motivated various efforts to implement OPF in a \emph{distributed} fashion, relying on agent-to-agent communication, such as via consensus algorithms~\cite{zhang_optimal_2015} or dual decomposition~\cite{bolognani_distributed_2013,dallanese_distributed_2013,sulc_optimal_2014}. 
Recent work has considered the co-optimization of planning and operation by considering both traditional expansion measures (such upgrading transformers or cables) as well as real-time control through DES, utilizing a decision-making process that builds on an iterative AC optimal power flow method~\cite{karagiannopoulos_hybrid_2017}.
Lastly, advances in extremum-seeking showed that model-free optimization is possible with provable guarantees for convergence and convexity for a variety of distribution feeder objective functions over a broad range of power flows~\cite{arnold_model-free_2016}.
Despite the elegance of distributed solutions, the necessary communication infrastructure is still a steep investment at the scale of transforming distribution networks. 

More recently, data-driven approaches to active distribution networks have been proposed. Some efforts propose data-driven methods to robustify OPF against the inherent uncertainty in injections from DERs~\cite{zhang_data-driven_2015,mieth_data-driven_2018}. These methods do not address the challenge of determining parameters for local controllers.
In~\cite{xu_data-driven_2018}, linear mappings are determined in real-time to approximate the nonlinear relationship between voltage magnitudes and
nodal power injections. While useful in scenarios where topological changes occur, the approach as is does not incorporate the abilities of OPF to produce network-wide sophisticated behaviors and constraint satisfaction.
The idea to use machine learning to learn how to mimic OPF by learning local mappings based on OPF calculations was first proposed in~\cite{sondermeijer_regression-based_2016}, and further formalized with information theory in~\cite{dobbe_fully_2017}. 
A related approach uses kernels to learn nonlinear localized control policies based on a linearized OPF problem~\cite{xu_data-driven_2018}.
More recently, the approach was applied to design voltage-reactive power (volt-var) droop-curves in~\cite{bellizio_optimized_2017,karagiannopoulos_data-driven_2019-3}. These approaches constrain the parameter space of the machine learning model to produce a monotonically decreasing volt-var relationship. 
While these methods can also mimic OPF behavior in a decentralized fashion, their reliance on voltage (an \emph{endogenous} control variable) easily leads to \emph{closed-loop feedback} instability issues~\cite{farivar_equilibrium_2013}, for instance when there is a mismatch between the voltages used during training and the voltage experienced in operation when the operator adjusts the substation voltage or as a result of interactions between the local controllers themselves.
Some related papers consider the use of robust optimization to infer decentralized linear decision rules for volt-var~\cite{jabr_linear_2018} or affine policies for controlling both real and reactive power~\cite{lin_decentralized_2018}. The former effort adopts the nonlinear bus injection model, while the latter assumes linear branch flow equations and includes intertemporal dependencies of battery storage systems.
While linear and (piecewise-)affine control policies have been central in the emerging debate on active distribution networks, they do form a limit on the kind of behaviors that may be attained and the types of OPF problems that can be decentralized.

\subsection{Contributions and Outline}

We propose to decentralize OPF by using supervised learning to learn inverter controllers that predict, and thereby approximate, optimal inverter actions from local measurements.
The method is agnostic in the structure of the machine learning models used, and relies on \emph{exogenous} variables as features, giving rise to an \emph{open-loop feedforward} optimization approach.
The method bridges control techniques based on local information with those based on system-wide optimization, thereby making contributions that overcome respective limitations.
Firstly, our method yields close-to-optimal results for various OPF objectives, for both single-phase and unbalanced three-phase networks. 
The method respects constraints on voltage, equipment specifications and power capacity across all validation scenarios.
Secondly, the implementation of decentralized OPF is data-driven, requires no manual controller tuning, and needs little or no real-time communication. The approach does require for updated controller parameters to be sent from the central design location to each inverter.
Thirdly, the open-loop nature of the method allows it be operated alongside legacy control equipment, such as load tap changers and capacitor banks, and does not compete or interfere with operator actions or other autonomous systems that make real-time control decisions. 
The learned controllers' strict dependence on disturbance variables not affected by control actions circumvents stability issues.
The feedforward nature does have the limitation that the controllers cannot detect sudden changes in the endogenous state, due to e.g. operator actions or faults, and a separate effort deals with combining feedforward and feedback mechanisms to allow for this~\cite{karagiannopoulos_data-driven_2019-3}.


The approach consists of four main steps. First, for a specific network with a set of controllable DERs, we gather historical data for loads and generators, collected over an extended period of time, typically collected by advanced metering infrastructure (AMI) or from simulations. For all scenarios, we run a centralized optimal power flow computation to understand how a group of DERs best minimizes a collective objective by adjusting real and/or reactive power injections, given certain capacity, operational and safety constraints (Section~\ref{sec:opf}).  
Secondly, we apply an information-theoretic framework to understand to what extent the local data of each DER allows \emph{any} parameterized controller to reconstruct the optimal set points and whether communication with other nodes is needed (Section~\ref{sec:ratedistortion}).
Thirdly, for each individual DER, we use supervised learning to determine a function that relates its \emph{local} measurements to its optimal power injections, as determined by OPF (Section~\ref{sec:regression}). 
Lastly, we implement these functions as controllers on the system to determine power injections based on a new local measurement that collectively mimic a centralized OPF scheme, which we validate on different test setups in single- and three-phase power flow (Section~\ref{sec:results}). 
The paper ends with a discussion about the effects of communication, the interpretation of the learned controller parameters and the challenges related to designing, planning and maintaining the learned controllers (Section~\ref{sec:discussion}).

\section{Optimal Power Flow}
\label{sec:opf}

We first set up the OPF framework, which enables us to compute the optimal set point for a collection of controllable DERs.
While we consider both singe- and three-phase networks, for the sake of brevity, we provide the derivation of OPF for single-phase settings only, directing the reader to three-phase OPF formulations in Section~\ref{sec:results_3ph}.
Adopting the full nonlinear \emph{DistFlow} model~\cite{baran_optimal_1989} for single-phase networks, we model a network $\G := (\N,\E)$ with buses or nodes $\N$ and edges or branches $\E$. The branch flow equations are
\begin{IEEEeqnarray}{Rl}
P_{mn} =& r_{mn} \ell_{mn} + p_{n}^\text{c} - p_{n}^\text{g} + u_n^p + \sum_{(n,k) \in \E, k \neq m} P_{nk} , \IEEEyesnumber\label{eqn:BFeqs1}\\
Q_{mn} =& \chi_{mn} \ell_{mn} + q_{n}^\text{c} - q_{n}^\text{g} + u_n^q + \sum_{(n,k) \in \E, k \neq m} Q_{nk}  , \IEEEyesnumber \label{eqn:BFeqs2} \\
y_{m} =& y_{n} + 2 \left( r_{mn} P_{mn} + \chi_{mn} Q_{mn} \right) \nonumber \\
& - \left( r_{mn}^2 + \chi_{mn}^2 \right) \ell_{mn} \,, \IEEEyesnumber \label{eqn:BFeqs3} \\
\ell_{mn} =& \frac{P_{mn}^2 + Q_{mn}^2}{y_m} \IEEEyesnumber\label{eqn:CurEq} \ , \ \forall (m,n) \in \E \,,
\end{IEEEeqnarray}
where $P_{mn},Q_{mn}$ denote the real and reactive power flowing out of node~$m$ towards node~$n$, $p_{n}^\text{c}$ and $q_{n}^\text{c}$ are the uncontrollable real and reactive power consumption at node~$n$, $p_{n}^\text{g}$ and $q_{n}^\text{g}$ are the uncontrollable real and reactive power generation at node~$n$, $y_n := V_n^2$ is the squared voltage magnitude, and $\ell_{mn} := I_{mn}^2$ is the squared current magnitude on branch~$(m,n)$.
We assume we are given a subset $\C \subset \N$ of buses that are equipped with a controllable source or load.
$u_n^p$ and $u_n^q$, for $n \in \mathcal C$, are the controllable parts of the real and reactive nodal power.
$r_{mn}$ and $\chi_{mn}$ are the resistance and reactance of branch~$(m,n)$. Here, we assume the nodal power consumption to be constant and not sensitive to voltage or current fluctuations.
This assumption is justified as loads are increasingly behaving as constant power due to many devices controlling their power intake or injection. 
Extension to include voltage sensitivity is possible but omitted here. For a formulation, see~\cite{farivar_inverter_2011}.

\subsection{Optimal Power Flow Objectives and Constraints}
We consider different controller architectures that can deliver real and reactive power $u_n^p,u_n^q$. 
Whether these are available depends on the controllable parts of the inverter-interfaced system, which may include both controllable generation, storage or load. 
%
We consider the following optimal power flow formulation:
\begin{IEEEeqnarray}{Rl}
\min_{\pmb{u}} \hspace{8pt} & f_o (\pmb{x},\pmb{u}) 
\IEEEyesnumber \label{eqn:OPF} \\
\text{s.t.} \hspace{8pt} & \eqref{eqn:BFeqs1}-\eqref{eqn:BFeqs3}, \nonumber \\
& \ell_{mn} \ge \frac{P_{mn}^2 + Q_{mn}^2}{y_m} \,, \forall (m,n) \in \E \,, \label{eqn:ConvRel} \\
& \text{capacity constraints on } u^p_i , u^q_i \,, \nonumber \\ 
& \underline{y} \leq y_{n} \leq \overline{y} \,, \forall n \in \N \,, \label{eqn:VoltConst}
\\
\pmb{x} =& \left(y_n, \ell_{mn} , P_{mn}, Q_{mn} \right),\, \pmb{u} = \left( u^p_i , u^q_i \right) \,, \nonumber \\
&\forall n \in \N \ , \ \forall (m,n) \in \E \ , \ \forall i \in \C \,. \nonumber
\end{IEEEeqnarray}
%
In our case studies in Section~\ref{sec:results}, we optimize various objectives relevant in distribution grid operations.
\begin{align}
 f_{o} :=& \ \alpha \displaystyle \sum_{(m,n) \in \E} r_{mn}\ell_{mn} + \beta \sum_{n \in \mathcal N} \left( y_n - y_{\text{ref}}  \right)^2 \nonumber \\ 
 &+ \gamma   \left( P_{01}^2 + Q_{01}^2  \right) \,.
   \label{eqn:ObjFunOp}
\end{align}
Firstly, to ensure the solution lies on the feasibility boundary of the inequality constraint~\eqref{eqn:ConvRel}, the convex relaxation requires an objective strictly increasing in the squared current magnitude~$\ell_{mn}$ and no upper bounds on the loads, thereby satisfying the original equality constraint~\eqref{eqn:CurEq}, as explained in~\cite{farivar_branch_2013}. 
We also use this objective to minimize real power losses.
Secondly, we minimize voltage deviations from a reference and voltage variability due to the intermittent nature of generation and consumption.
This objective aims to actively keep the voltage in a safe and desirable operable regime.
Lastly, we aim to make a network \emph{self-sufficient} by minimizing power delivered from the transmission grid, injected at the top of a radial network, thereby maximizing the use of distributed energy resources to provide loads nearby. 
$\alpha, \beta,\gamma$ denote hyperparameters, and $y_{\text{ref}}$ the reference voltage throughout the network.
In this paper, we are primarily concerned with attaining certain qualitative behaviors with OPF. If one has an economic consideration, the hyper parameters can be scaled to reflect the economic cost of various objectives.
The nonlinear equality constraint \eqref{eqn:CurEq} is relaxed to \eqref{eqn:ConvRel}, enabling a convex second order cone program~\cite{farivar_branch_2013}. 
Equation~\eqref{eqn:VoltConst} formulates the responsibility of American utilities to maintain service voltage within $\pm 5\%$ of 120$V$ as specified by ANSI Standard C84.1.

Two constraints are included to account for inverter capacity and voltage goals. 
Each controller $i \in \C$ is ultimately limited by a local capacity on total apparent power capacity $\bar{s}_i$, which depends on the other functions an inverter executes. 
Most inverters are able to provide real and reactive power in a coupled fashion through a \emph{four-quadrant} configuration, which yields a disk constraint,
\begin{equation}
(u_{i}^p)^2(t) + (u_{i}^q)^2(t) \le \bar{s}_i^2(t) \,.
\label{eqn:InvCap1}
\end{equation}
In some cases, capacity may be assigned in a decoupled fashion for real and/or reactive power, in which an inner approximation of the disk constraint leads to
\begin{equation}
\underline{p}_i(t) \le u_{i}^p(t) \le \overline{p}_i(t) \ , \ \underline{q}_i(t) \le u_{i}^q(t) \le \overline{q}_i(t) \,,
\label{eqn:InvCap}
\end{equation}
with $\underline{p}_i,\overline{p}_i$ denoting the lower and upper capacity constraints on nodal power injection.
Lastly, we consider inverters that also interface a system with a local photovoltaic (PV) installation, an electric vehicle (EV) or battery charging system, smart loads or a combination thereof.
While incorporating the temporal capacity of batteries and its intertemporal dependencies as a constraint is out-of-scope for this work, we can indirectly model time-varying capacity when historical data of capacity is available or can be computed.
Due to the nature of these subsystems having some hierarchy of priority, it can happen that the capacity available for the OPF problem is dependent on the scheduling and activity of these other systems.
In this paper, we consider a case where inverters can only deliver or consume reactive power based on the remaining capacity \emph{after} injecting surplus energy from the PV installation into the grid. 
In this case, the reactive power capacity $\bar{q}_i(t)$ at time $t$ of an inverter is limited by the total apparent power capacity $\bar{s}$ (constant) minus the real power generation $p_i^{\text{g}}(t)$. 
As a result, the demand of reactive power does not interfere with real power generation, which is formulated as
\begin{equation}
\left| u^q_i(t) \right|  \leq \bar{q}_i (t) = \sqrt{\bar{s}_i^2 - (p_i^{\text{g}}(t))^2} \,.
\label{eqn:InvCap2}
\end{equation}
We will use this constraint in case 1, where we assume that each inverter has some overcapacity with respect to the maximum real power output, e.g. $\bar{s}_i = 1.05 \bar{p}_i$.



\subsection{A Principled Partitioning of Variables}

To formalize the reconstruction, we first reformulate the state variables per bus as
\begin{equation}
x_i := 
\left[
\begin{matrix}
V_i \\
\delta_i \\
p_i \\
q_i \\
\end{matrix} 
\right] 
\in \R^{4} \,.
\label{eqn:state}
\end{equation}
Based on the bus type, we can partition the state~$x$ into controllable inputs~$u$, uncontrollable inputs or disturbances~$d$ and endogenous variables~$x^{\text{end}}$, as suggested in~\cite{hauswirth_online_2017} and summarized in Table~\ref{tab:bus_types} for basic bus types. Note that an actual bus may be a combination of basic bus types (such as combining load and generation, or a controllable load), and $p_i,q_i$ denote the net nodal power. 

\begin{table}[h]
\centering
\caption{Partitioning of variables for all bus types, as proposed in~\cite{hauswirth_online_2017}.}
\begin{tabular}{lccc} 
\toprule
 				& \multicolumn{2}{c}{Exogenous} 			& Endogenous 		\\ \cmidrule(r){2-3}
				& Controlled - $u$	& Uncontrolled - $d$ 	& 	$x^{\text{end}} := x_{\backslash \{ u,d \} }$			\\ \midrule
PQ generation		& $p_n,q_n$		&					& $V_n,\delta_n$	\\
PQ load			& 				& $p_n,q_n$			& $V_n,\delta_n$	\\
PV generation		& $p_n,V_n$		& 					& $q_n, \delta_n$	\\
slack bus			& $V_0$			& $\delta_0$			& $p_0,q_0$		\\ \bottomrule
\end{tabular}
\label{tab:bus_types}
\end{table}

An important reason to distinguish between endogenous and exogenous variables, is that these have a different effect on the dynamics when chosen as controller variables. Endogenous variables implicate closed-loop feedback control, while exogenous variables lead to an open-loop feedforward control scheme.
In this paper we focus our attention on the latter, using exogenous disturbances $d_i$ as our controller variables. 
This choice circumvents the potential for instability inherent to feedback policies, such as those known to exist for volt-var control schemes~\cite{farivar_equilibrium_2013}. 
In addition to exogenous variables directly impacting the optimal power flow problem, as listed in Table~\ref{tab:bus_types}, we consider other exogenous variables $d^{\text{other}}$ that may help explain our data, such as time and weather variables that are readily available.
In a separate effort, we compare open-loop and closed-loop configurations~\cite{karagiannopoulos_data-driven_2019}.

\subsection{Building A Data Set Of Optimal Set Points}
We treat the $i^{\text{th}}$ agent's optimal action in the centralized problem as a random variable $u_i^*$, which we define as an element in the set of minimizers of problem~\eqref{eqn:OPF}-\eqref{eqn:VoltConst}, and which depends on the global state variables $x = (x_1,\hdots,x_{|\N|})$. 
In the OPF context, the goal of the machine learning procedure is to find a model for each DER $i \in \C$ that approximates the optimal power injection $u^*_i$ based \emph{solely} on local measurements~$d_i$ available at node $i$.
We construct a \emph{central training set} consisting of power flow scenarios that are representative of future behavior on the grid. This data set can contain historical measurements taken from advanced metering infrastructure (AMI) in the grid, augmented with proxy data for variables that were not measured or for scenarios that did not occur in the past, but are anticipated to occur in the future. For example, a load not metered by AMI, may be approximated with an average load profile. 
Examples of anticipated future behaviors not captured in historical measurements are connections of new electric vehicles or solar installations. 
If known a priori, these may be simulated and added to the training set.
Together, the central training set contains $T$ power flow scenarios augmented with the optimal control set points $\{x(t), u^*(t)\}_{t=1}^T$, which were computed in a centralized OPF problem as defined in Section~\ref{sec:opf}.

\section{Assessing Data For Local Reconstruction}
\label{sec:ratedistortion}

The core challenge we are facing is developing a predictor $\hat u_i$ that can \emph{reconstruct} the optimal action $u_i^*$ (which depends on the whole network state $x$), based on solely local information $d_i$ (a subset of $x_i$ as defined in Equation \eqref{eqn:state}). 
Traditional machine learning methods do not provide a straight-forward way to assess how well a predictor performs in an absolute sense.
To provide such a measure, we formulate our approach as a \emph{data compression} problem.
Analogue to reducing the number of pixels in a JPG photo, we reduce the number of states that a predictor has access to, from all states $x$ (that fully describe the optimal action $u_i^*$) to the local exogenous state variables $d_i$ (that inherently only hold limited information about $u_i^*$).

\begin{figure*}[t]
\centering
  \includegraphics[width=0.75\textwidth]{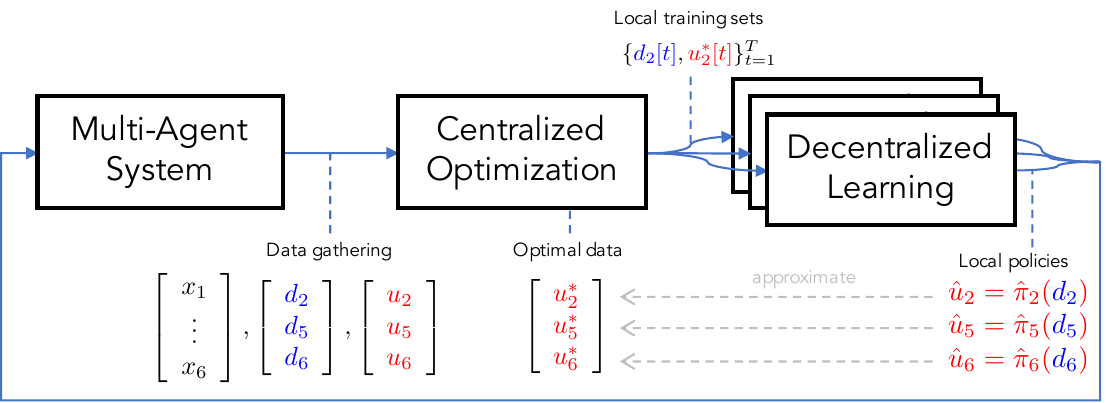}
  \caption{A flow diagram explaining the key steps of the decentralized regression method. We first collect data from a multi-agent system, and then solve the centralized optimization problem using all the data. The data is then split into smaller training and test sets for all agents to develop individual decentralized policies $\hat \pi_i(x_i)$ (or $\hat \pi_i(d_i)$ as proposed in the context of OPF) that approximate the optimal solution of the centralized problem. These policies are then implemented in the multi-agent system to collectively achieve a common global behavior.}
  \label{fig:flow_diagram}
\end{figure*}

\subsection{Rate Distortion Framework}

With \emph{rate distortion} theory~\cite[Chapter 10]{cover_elements_2012}, a branch of information theory, we cast this problem formally to understand the inherent distortion that comes with compressing the data produced by an information source into a specified \emph{encoding rate}. In our case, this rate is equivalent to the \emph{mutual information} $I(d_i,u_i^*)$ between accessible variables $d_i$ and the optimal action $u_i^*$, both treated as random variables. 
For our problem, the theoretical minimum distortion given the encoding rate is
\begin{align}
\label{eqn:rate_distortion_formulation}
D^* = \min_{p(\hat u | u^*)} \, & \quad \Ebb \left[\pmb{d}(\hat u, u^*) \right] \,, \\
\textnormal{s.t. } \, & \quad I(\hat u_i; u_j^*) \le I(d_i; u_j^*) \triangleq \gamma_{ij} \,, \nonumber\\
& \quad I(\hat u_i; \hat u_j) \le I(d_i; x_j) \triangleq \delta_{ij}, \forall i, j \in \C \,, \nonumber
\end{align}
%
%
%
where $I(\cdot,\cdot)$ denotes mutual information and $\pmb{d}(\cdot,\cdot)$ an arbitrary non-negative distortion measure. 
For the remainder of this paper, we will measure distortion it as the mean square error (MSE) deviation between $\hat u_i$ and $u_i^*$, which is a common \emph{loss function} in supervised learning. 
The minimum distortion between random variable $u^*$ and its reconstruction $\hat u$ hold over \emph{any} conditional distributions $p(\hat u | u^*)$.
Note that this result holds for any non-negative distortion measure. 

Solving for all conditional distributions $p(\hat u | u^*)$ is not feasible nor is it necessary for our purpose.
We are strictly interested in finding the best parameterized Markov control policy $\hat u_i = \hat \pi_i(d_i)$ among a \emph{finite set} of available supervised learning techniques.
As such, we do not need to compute $D^*$, and use \eqref{eqn:rate_distortion_formulation} to provide us with two practical tools to assess the quality of the data set. 

Firstly, regardless of the distortion measure~$\pmb{d}(\cdot,\cdot)$, minimum distortion~$D^*$ is monotonically decreasing in the encoding rate $I(d_i; u_j^*)$.
Equivalently, policies $\hat u_i$ that have a higher mutual information $I(\hat u_i; u_i^*)$ also yield lower distortion rates, with the minimum occurring for $I(\hat u_i; u_i^*) = I(d_i; u_i^*)$. 
Therefore, we can use $I(d_i; u_i^*)$ as a benchmark for assessing various policies. 
Mutual information cannot be computed exactly but can be approximated numerically. To do this, the data sets for all three quantities, $\{\hat u_i(t)\}_{t=1}^T, \{ u_i^*(t)\}_{t=1}^T, \{ d_i(t)\}_{t=1}^T$, are discretized into even buckets, allowing the use of an estimation algorithm to compute mutual information~\cite{jiao_minimax_2015}.
Note that a finer discretization yields a better estimate, but also requires more memory. Depending on the available computational resources, a feasible tradeoff should be made. Explicit results of these computations are included in~\cite{dobbe_fully_2017}.


Secondly, if the policy that attains the highest possible mutual information does not provide good performance, the rate distortion framework provides a ready extension to determine the optimal nodes to communicate with, thereby increasing the mutual information benchmark (and decreasing the minimally achievable distortion~$D^*$), which is discussed next. 



\subsection{Allowing Restricted Communication}
Suppose that a decentralized policy $\hat \pi_i (d_i)$ suffers from the fact that the data set with local measurements $\{ d_i[t] \}_{t=1}^T$ does not contain enough information to make an accurate reconstruction of the optimal action $u^*_i$ that generalizes well across power flow scenarios. 
An example of such a scenario is when the optimal set points do not correlate with any local process, instead responding more to behaviors elsewhere in the network. We simulate such an example in Section~\ref{sec:results_case2} and discuss it further in Section~\ref{subsec:communication}.
Such a deficiency in local information can be determined by estimating the mutual information $I(d_i,u_i^*)$ or by inspecting the prediction accuracy of resulting machine learning models.
Since $u^*_i$ depends on all states in the network $x$, we would like to quantify the potential benefits of communicating with other nodes $j \neq i$ in order to reduce the distortion limit $D^*$ from \eqref{eqn:rate_distortion_formulation} and thereby improving the reconstruction $\hat u_i = \hat \pi_i (d_i,d_j)$ as compared to $\hat u_i = \hat \pi_i(d_i)$.
\begin{lemma} (Restricted Communication~\cite{dobbe_fully_2017})
\label{thm:restricted_communication}

If $\Sc_i$ is the set of $k$ nodes $j \ne i \in \N$ which $\hat u_i$ is allowed to observe in addition to $d_i$, then setting 
\begin{align}
\label{eqn:restricted_communication}
\Sc_i = \arg \max_{\Sc} \ I(u_i^*; d_i, \{d_j : j \in \Sc\}) \ : \ |\Sc| = k \,,
\end{align}
minimizes the best-case expectation of \textit{any} distortion measure. That is, this choice of $\Sc_i$ yields the smallest lower bound $D^*$ from \eqref{eqn:rate_distortion_formulation} of any possible choice of $\Sc$.
\end{lemma}

Lemma \ref{thm:restricted_communication} provides a means of choosing a subset of the state $\{d_j : j \ne i\}$ to communicate to policy $\hat \pi_i$ that minimizes the corresponding best expected distortion $D^*$. Practically speaking, this result may be interpreted as formalizing the following intuition: ``the best thing to do is to transmit the most information.'' In this case, ``transmitting the most information'' corresponds to allowing $\hat \pi_i$ to observe the set $\Sc$ of nodes $\{d_j : j \ne i\}$ which contains the most additional information about~$u_i^*$. Likewise, by ``best'' we mean that $\Sc_i$ minimizes the best-case expected distortion~$D^*$, for any distortion metric~$\pmb{d}$. 
Following the logic above of an example where there is little local information about $u_i^*$, the resulting nodes $\Sc_i$ are likely to exhibit behaviors that $u^*_i$ responds to in the considered OPF problem.
We determine $\Sc_i$ by evaluating the mutual information for all possible sets $\{d_j : j \in \Sc\}) \ : \ |\Sc| = k$.
In the case of selecting multiple nodes to communicate with, one may encounter the estimation to require significant memory and slow down. A greedy approach, selecting one node at a time, may prove useful, and has near-optimal properties due to the submodularity of mutual information~\cite{sharma_greedy_2015}.
%

\section{Machine Learning}
\label{sec:regression}

Regression is now performed for each individual DER, by selecting a \emph{local data set} from the central data set. As discussed, we deliberately choose to only select exogenous uncontrollable variables $d_i$ that are measured at node~$i$. As such, we do not consider states in $x_i$ that are dynamically coupled with the control $u_i$, thereby circumventing instability arising from controller interactions. The variables contained in~$d_i$ may differ from node to node based on the local sensing infrastructure. These can be complemented by other predictive variables that are available locally such as time or weather information~$d^{\text{other}}$.
For illustration, we use a running example of a node that can measure only the local load~$p_i^{\text{c}}$, the local PV generation~$p_i^{\text{g}}$, and the present capacity~$\overline{s}_i$. This yields the local training set with base variables and labels,
\begin{equation}
     X_i = \left[ d_i(1) \cdots d_i(T)  \right] \,, \, Y_i = \left[ u^*_i(1) \cdots u^*_i(T) \right] \,.
\end{equation}
The next step is to select a machine learning model that is expressive enough to reconstruct valid patterns in the optimal control actions that can be found in the local data, and that is simple enough to generalize well to new scenarios.
We compared various model structures in our earlier work~\cite{sondermeijer_regression-based_2016} and selected a multiple stepwise linear regression algorithm that selects a subset of available nonlinear features~\cite[Section 3.3]{friedman_elements_2001}. 
This model is linear in the parameters, but allows for nonlinear transformations of the base variables in $\varphi(d_i)$, concretely quadratic and interaction terms.
For instance, if~$d_i$ contains 3 base variables as in our example, $\varphi(d_i)$ would yield 3 linear features; $\varphi_1^{(i)} := p_i^{\text{c}}$, $\varphi_2^{(i)} := p_i^{\text{g}}$ and $\varphi_3^{(i)} := \overline{s}_i$, and 6 more nonlinear features, of which 3 interaction terms ($\varphi_4^{(i)} := \varphi_1^{(i)} \varphi_2^{(i)}$ etc.) and 3 quadratic action terms ($\varphi_7^{(i)} := (\varphi_1^{(i)})^2$ etc.).
In general, for $B_i$ base variables, we construct $F_i = 2B_i+ \binom{B_i}{2}$ features as inputs to the regression problem.
The $F_i$ input variables of the $t^{\text{th}}$ sample are denoted as $\pmb{\phi}^{(i)}(t) \in \R^{F_i}$.
All $F_i$ features across all $T$ data points can be organized as
\begin{equation}
\Phi^{(i)} = \begin{bmatrix}
\pmb{\phi}^{(i)\top}(1) \\
\vdots \\
\pmb{\phi}^{(i)\top}(T) 
\end{bmatrix} = 
\begin{bmatrix}
\varphi_1^{(i)}(1) & \hdots & \varphi_{F_i}^{(i)}(1) \\
\vdots & \ddots & \vdots \\
\varphi_1^{(i)}(T) & \hdots & \varphi_{F_i}^{(i)}(T)
\end{bmatrix} \,. \label{eq:RegX}
\end{equation}
We use a linear model to relate output $Y_i$ to input matrix $\Phi^{(i)}$,
\begin{equation}
 \hat{\pi}_i(\beta^{(i)},\pmb{\phi}^{(i)}(t)) = \beta^{(i)}_0 + \beta^{(i)}_1 \varphi^{(i)}_{1}(t) + \hdots + \beta^{(i)}_{F_i} \varphi^{(i)}_{F_i}(t) \,. \label{eq:MLmodel}
\end{equation}
A least squares approach determines the coefficients ${\beta^{(i)}} = \left[ \beta^{(i)}_0, ..., \beta^{(i)}_F \right]$ that minimize the residuals sum of squares \eqref{eq:RSS} given $T$ samples,
\begin{align}
\text{RSS}( \beta^{(i)} ) &= \sum^T_{t=1} \left(u^*_i(t) - \hat{\pi} \left({\beta^{(i)}}, \pmb{\phi}^{(i)}(t) \right) \right)^2 \,. \label{eq:RSS}
\end{align}
\noindent The algorithm is initialized with a multiple linear model of the basic variables in $X_i$ only. At each iteration, the new feature that improves the Bayesian Information Criterion (BIC) \cite{schwarz_estimating_1978} the most and sufficiently is added to the model. Subsequently, the variable with the lowest contribution is removed. These two steps are iterated until no variables meet the entrance or exit threshold of the algorithm. The goal of the stepwise selection algorithm is to select the subset of features that most accurately predicts the optimal power injection of a DER.

\section{Results}
\label{sec:results}

We evaluate the proposed method on a realistic testbed that is constructed from two independent sources: we construct a 129 node feeder model based on a real distribution feeder from Arizona (Figure \ref{fig:FeederDiag}). Pecan Street power consumption and PV generation data with a resolution of 15 minutes is obtained from 126 individual residences in Austin, Texas for a period of 330 days, starting on January 1, 2015~\cite{noauthor_dataport_2017}.
Individual household load and PV time series are selected randomly from the Pecan Street data set and aggregated to match the spot load for each bus as specified in the experiment.
We demonstrate Decentralized OPF with three experiments.
The first experiment assumes controllable DER and loads spread randomly across a network. In this experiment, only reactive power is controlled with the objective to minimize voltage variations throughout the network.
The second experiment considers a group of controllable DER concentrated in one area, and a group of large loads concentrated in another area. In this experiment, both real and reactive power are controlled and the objective is to produce power locally as much as possible.
The third experiment considers Decentralized OPF in a three-phase setting, where linear approximations are used in OPF to enable balancing of voltage across phases.

\subsection*{Case 1: Minimize Losses and Voltage Variations}
For the first experiment, 27 of the 53 nodes with loads are randomly selected and equipped with PV installations with peak generation 80$\%$ of the peak real power load.
PV data is retrieved for different households in the Pecan Street dataset~\cite{noauthor_dataport_2017}, and aggregated to match peak generation.
We solve 2500 instances of OPF with sampled load and PV generation data to retrieve the optimal reactive power output of all inverters~$\{u^*[t]\}_{t=1}^T$, with the objective set with~$\alpha=1,\beta=2 \cdot 10^{-4}, \gamma = 0$. 
We consider the control of reactive power constrained by real power generation as formulated in~\eqref{eqn:InvCap2}. 
The data is divided into training, test and validation data. Test data is used to test the predictive performance of the machine learning models. The models are then applied in simulation with validation data. 
\begin{figure}[!t]
\centering
\includegraphics[width=.48\textwidth]{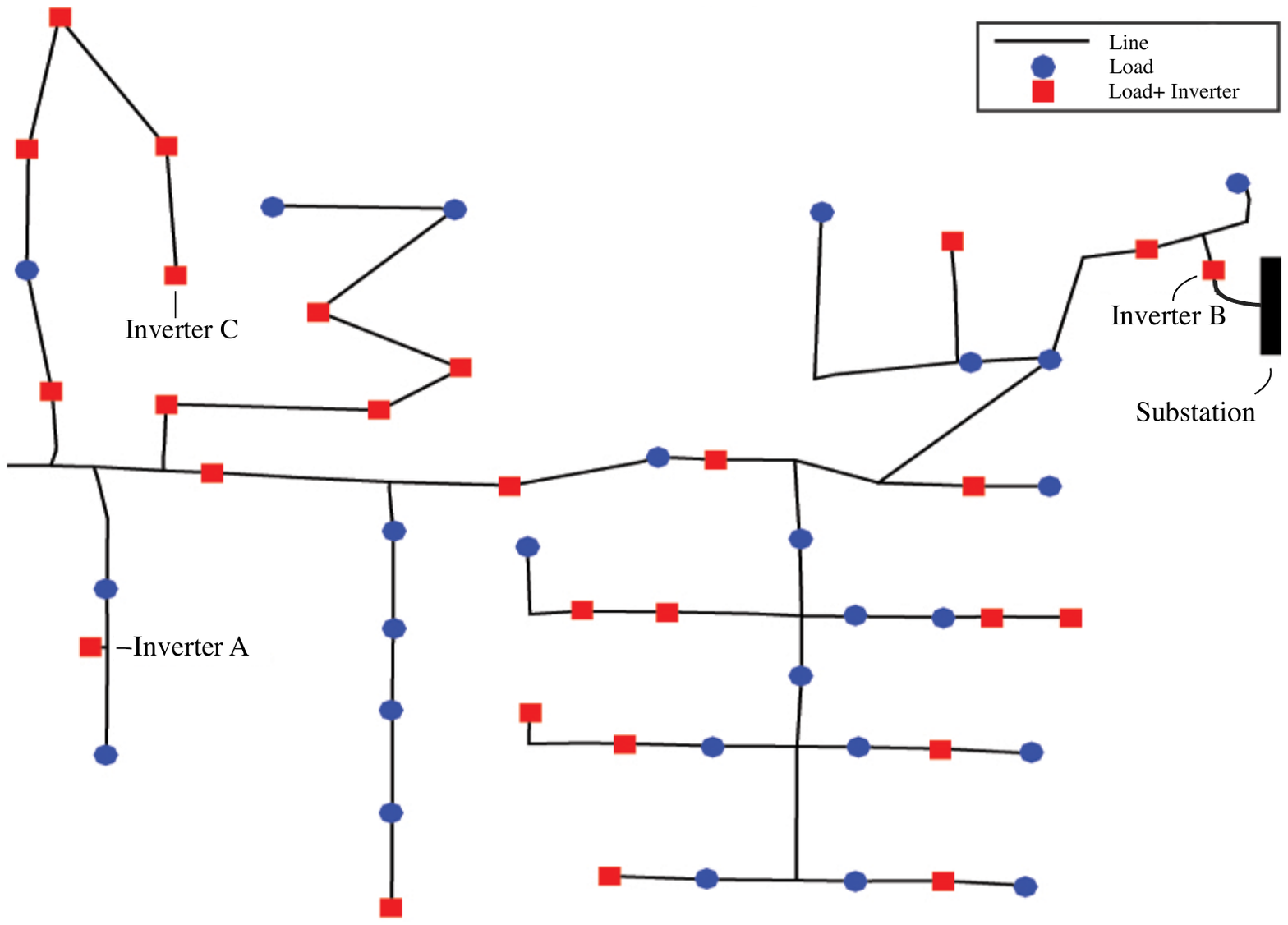}
\caption{Distribution feeder topology for case 1 and 2, with load and inverter assignment for case 1. Substation is located on the far right, locations of loads with PV inverters (squares), and without PV inverters (circles) are included.}
\label{fig:FeederDiag}
\end{figure}
%
%
%
%

The data-driven decentralized OPF approach is simulated and compared to two other scenarios: a situation where inverter reactive power capacities are not utilized, and a scenario where inverters are operated around a constant, non-unity, power factor, as proposed in the German grid code~\cite{vde-ar-n_4105_power_2011}.
In our context, we chose to tune the inverters to operate at lagging (generating) power factor of 0.9 to reduce losses. 
In addition to the comparison, we extend our method to show it is capable of collaboration with a load tap changer (LTC). 
We design a scheme in which the inverters operate with controllers to flatten the voltage throughout the feeder, and then adjust the turn ratio of the LTC at the substation, to safely lower the voltages throughout the feeder, without violating the lower voltage constraints.
As such, we consider four approaches: a) no reactive power support, b) constant power factor support, c) data-driven decentralized OPF, and d) data-driven decentralized OPF collaborating with substation load tap changer.

\begin{figure}[!t]
\centering
\includegraphics[width=.48\textwidth]{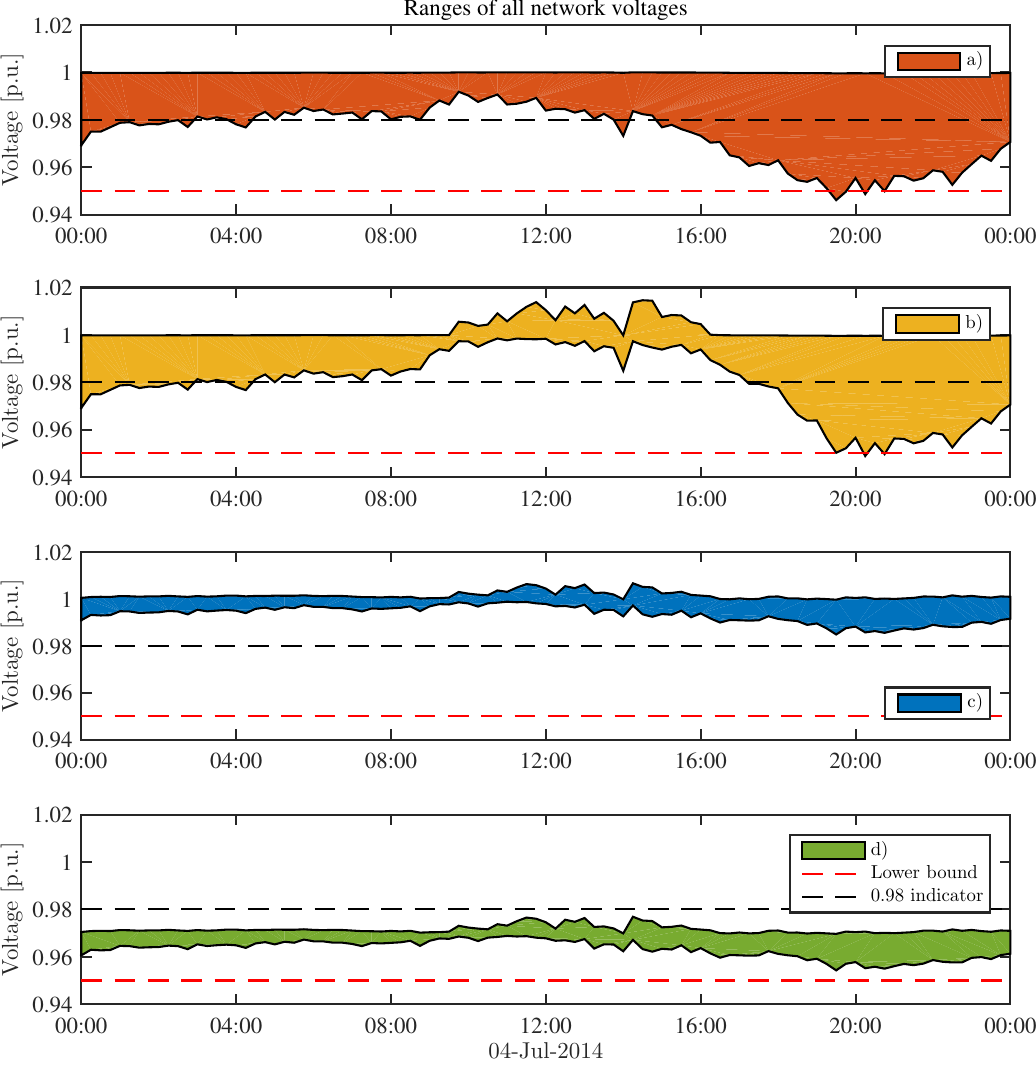}
\caption{Range of network voltages for the case of no control and applying the regression-based Decentralized OPF method. Colored planes represent the range between the maximum and minimum voltages across all network buses. Lower voltage bound is indicated with red dashed line.}
\label{fig:VoltSim}
\end{figure}

\begin{figure}
\centering
\includegraphics[width=.48\textwidth]{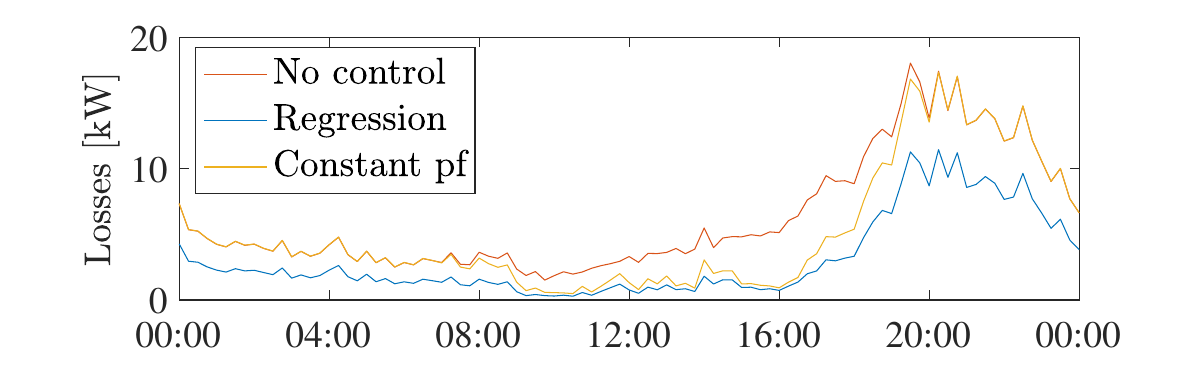}
\caption{Comparison of losses with no control and constant power factor method.}
\label{fig:LossComparison}
\end{figure}

Figure~\ref{fig:VoltSim} shows the voltage ranges across a full day for the different approaches.
It is relevant to note, that here we do not plot the results for OPF, as these are indistinguishable from the Decentralized OPF method.
For the case of no control, the voltage drop in the system is smallest between 10:00--16:00, when most real power demand is supplied by PV systems. 
The data-driven decentralized OPF method achieves system voltages that are close to the nominal value of 1 p.u., and simultaneously reduces losses as shown in Figure~\ref{fig:LossComparison}. 
The transition from peak PV generation to peak consumption between 12.00 and 20.00 causes the system voltage to change significantly without control. 
For the case of no control, the lower bound of the ANSI standard is violated in the evening if traditional voltage regulators are not operated. 
The effect of Decentralized OPF is obvious at these times: reactive power generated by inverters minimizes voltage drops in the system and reduces losses.

Figure \ref{fig:LossComparison} compares the losses for approaches a), b), and c). Compared to the situation of approach a), both approach b) and c) have beneficial effect on the objective function. However, approach c) achieves the best performance across time. Approach b) generates reactive power proportional to real power output, which is reflected in a lack of control during hours when PV power is not available.
Between 12.00 and 20.00 the objective function value of approach a) increases rapidly, which is caused by the transition from peak real power generation to peak real consumption. The objective function value of approach c) also increases, but significantly less. 

Compared to the objective values of the original OPF problem, the data-driven decentralized OPF method performs near-optimal at a difference of $0.15\%$ on average, with a maximum of 1.6$\%$ as compared to the objective value achieved by OPF.  
The bottom graph in Figure~\ref{fig:VoltSim} shows the resulting voltage range for the scenario where the LTC is adjusted such that the voltage at the substation is set at 0.97 p.u.. We see that the decentralized controllers collectively maintain the ability to flatten the voltage throughout the network and keep the voltage above the lower constraint boundary.

\begin{figure}[!t]
\centering
\includegraphics[width=.48\textwidth]{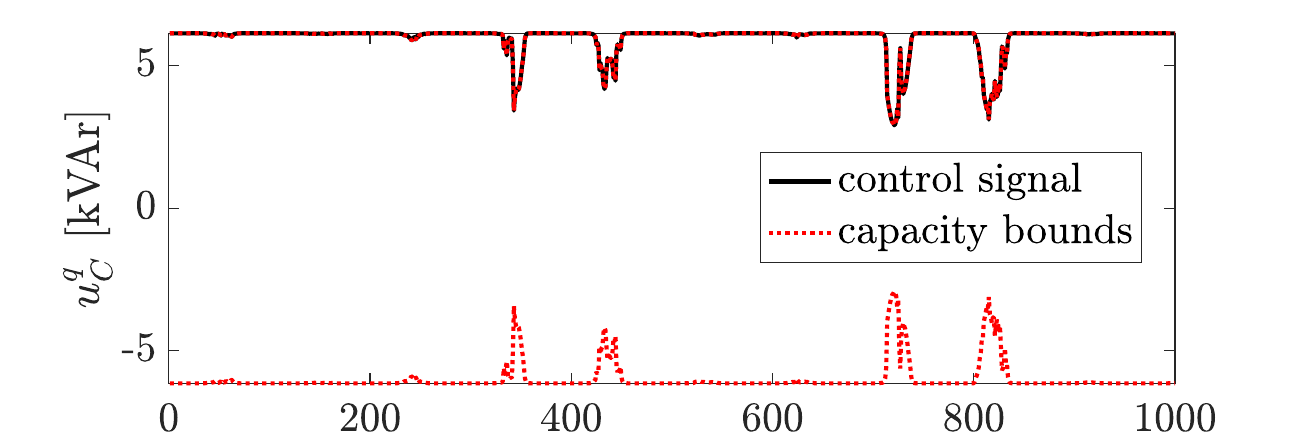}
\caption{Representative example of a control signal for inverter C (in Figure~\ref{fig:FeederDiag}). Notice that the inverter is operating at full capacity all the time, and has learned to respect its upper capacity boundary.}
\label{fig:controlsignal}
\end{figure}

While the set of controllers produce desirable global behaviors and constraint satisfaction in the network, it is also relevant to see if individual controllers respect their local capacity constraints. While we eventually implement a saturation rule that prevents the signal from going beyond its capacity boundaries, we also analyze the percentage of constraint violations on the validation data. 
Figure~\ref{fig:controlsignal} shows the resulting control signal across all training and validation data for Inverter C (see Figure~\ref{fig:FeederDiag}). Notice that the controller performs at its upper capacity. It has learned to respect this constraint from the data itself.
For models trained with a training set of only 700 data points, violations happen on average 0.00025\% on the test data, with a maximum of 0.33\% over all nodes. It is important to realize that a controller may learn to do a better job at satisfying its own local constraints, if the training data includes sufficient cases where the inverter hits its constraint boundaries in the solution of the OPF problem.


\subsection*{Case 2: Localize Power Generation}
\label{sec:results_case2}
\begin{figure}[h]
    \centering
        \includegraphics[trim=1.2cm 0 1.2cm 0,width=.45\textwidth]{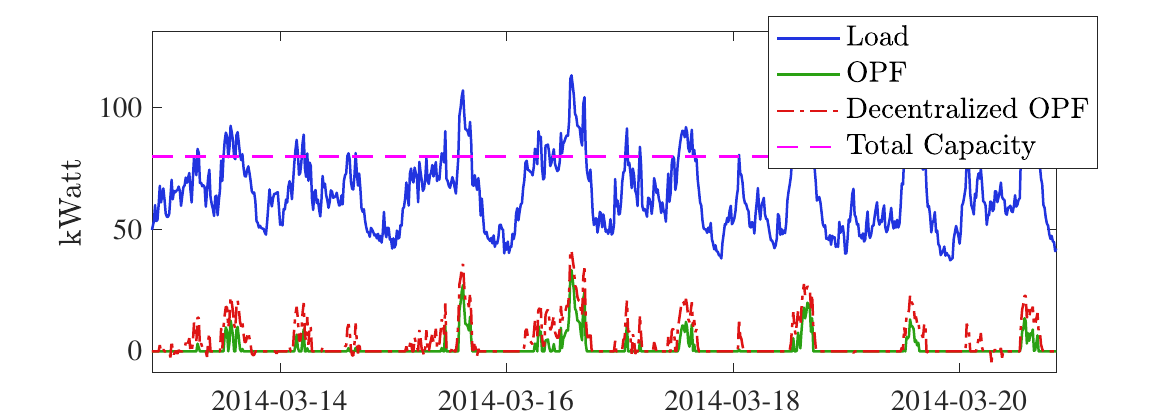}
    \caption{Total net load delivered by the substation, in the case of: no control, OPF and Decentralized OPF.}
    \label{fig:case2result}
\end{figure}
Case 2 aims to source power locally, thereby minimizing the reliance on the transmission grid.
We use the same feeder network topology and impedances, as depicted in  Figure~\ref{fig:FeederDiag}, with a different load and inverter assignment.
We assign 20 nodes in the lower right subradial with controllable DERs. 
In addition, the spot loads of 12 nodes in the two left upper subradials are increased by a factor 3. 
By setting parameter~$\gamma$ in the objective function sufficiently high (here we set $\alpha=1,\beta=0,\gamma=10^3$), this OPF problem tries minimize power procurement from the transmission grid by matching supply and demand locally, letting power flow from the area with controllable DER to the area with concentrated loads.
The focus is on controlling real power injections, with constraints formulated as in~\eqref{eqn:InvCap}.
The total real power capacity of the controllable DERs is assumed to be constant; future work will introduce time-varying capacity profiles and inter-temporal dependencies based on charging, generation and consumption patterns. 
We assign 4 kWatt capacity to 20 inverters, totalling to 80 kWatt generation/consumption capacity across the feeder, as indicated by the dotted red line in Figure~\ref{fig:case2result}.
Figure~\ref{fig:case2result} shows the approximate nature of Decentralized OPF as compared to the central OPF solution. That said, Decentralized OPF does a fine job at how mimicking the centralized OPF solutions across the test and validation data.
Notice how the power delivered by the substation is 0 for all times where the network's net load is smaller than the combined power capacity of 80 kWatt.



\subsection{Case 3: Phase Balancing in a Three-Phase System}
\label{sec:results_3ph}

\begin{figure}[h!]
\centering
\includegraphics[trim=0.5cm 0 1.5cm 0,width=.48\textwidth]{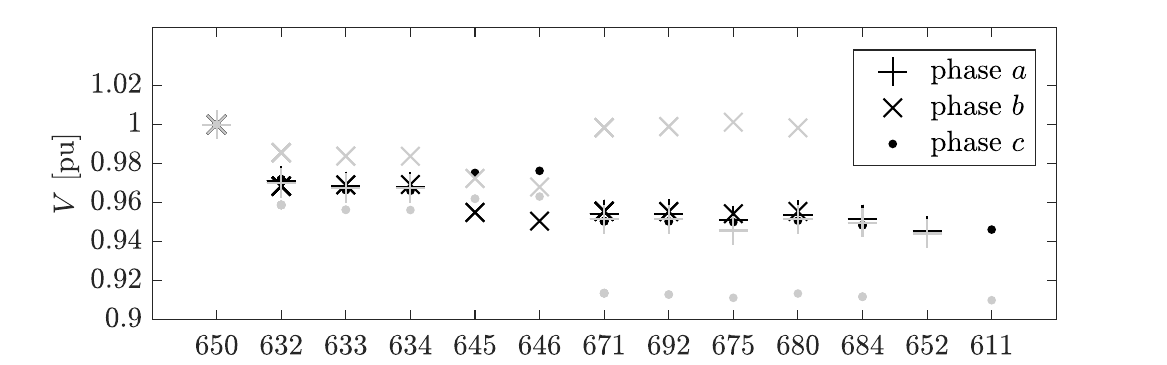}
\caption{Voltage magnitude balancing. In grey, voltage magnitudes across three phases without control. In black, voltage magnitudes across three phases applying decentralized OPF.}
\label{fig:Result3Phase}
\end{figure}

Whereas the use of convex relaxations is well-applicable in single-phase networks, these are not yet robust enough for the more complex OPF formulations for unbalanced three-phase systems. Its development for three-phase networks~\cite{dallanese_distributed_2013} is hampered by issues of nondegeneracy and inexactness of the semidefinite relaxations~\cite{louca_nondegeneracy_2014}. In~\cite{sankur_linearized_2016}, a linear approximation of the unbalanced power flow equations is proposed to overcome these challenges and enable three-phase OPF for a broad range of conditions.
Here we adopt the modeling developed in~\cite{arnold_optimal_2016, sankur_linearized_2016,sankur_optimal_2018}, which can be seen as an adaptation of the \emph{DistFlow} model~\cite{baran_optimal_1989} to unbalanced circuits, coined the \emph{Dist3Flow} model.
We show the efficacy of Decentralized OPF for three-phase unbalanced networks, by learning and reconstructing the solutions~$u_{n}^{p,\phi},u_{n}^{q,\phi}$, with phase $\phi \in \{ a,b,c\}$, of the centralized OPF problem~\cite[Eqn. 25]{sankur_linearized_2016}, with the main goal to minimize voltage differences across phases (the objective to be minimized includes the term~$\sum_{n \in \N} \sum_{\phi \neq \psi} (y_n^{\phi} - y_n^{\psi})$).
The feeder model used here is the IEEE 13 node test feeder~\cite{noauthor_ieee_2017}.
Figure~\ref{fig:Result3Phase} shows the results of applying Decentralized OPF to balance voltage magnitudes across phases for a scenario with a significant gap between phase~$b$ and phase~$c$ of almost $0.1$p.u..
We see that the voltage magnitudes are tightly balanced across the network. Notice that in order to balance the large gap in the lower half of the feeder (nodes 671, 692, 675, 680, 684), a small increase in gap is incurred in the left upper radial (nodes 645, 646).
Similar to the results for single-phase networks, the Decentralized OPF method also performs well at locally reconstructing OPF solutions in three-phase unbalanced networks, further validating its promise to serve as a new paradigm for regulating voltage and power flow in distribution networks.

\section{Discussion}
\label{sec:discussion}


In Section~\ref{subsec:communication}, we analyze how the optimal communication strategy, as determined via Lemma~\ref{thm:restricted_communication} and \eqref{eqn:restricted_communication}, differs between scenario 1 and 2. 
In Section~\ref{subsec:interpretation}, we interpret the resulting controller parameters for inverters located in different parts of the feeder.
Lastly, in Section~\ref{subsec:design}, we discuss practical considerations to integrate the decentralized OPF method across distribution planning and operation. 

\subsection{The Effect of Communication}
\label{subsec:communication}


The ability to fully decentralize OPF problems is surprising.
The information theoretic lens used in this paper complements insights from control-theoretic analysis that show the necessity of communication for a large class of local controllers to ensure that equilibria are feasible with respect to voltage constraints~\cite{cavraro_value_2016}. 
In our setting, the use of exogenous disturbance variables~$d_i$ prevents the need for a dynamic analysis as $d_i$ is not affected by the control signals.
Rather than for the purpose of stability and feasibility, in~\cite{dobbe_fully_2017}, we applied Lemma~\ref{thm:restricted_communication} to the OPF problem for a smaller network \cite{noauthor_ieee_2017}, in order to study the value of communication for improving the reconstruction of $u^*$, yielding an optimal communication strategy. 
%
%
In this paper, we also studied the emerging communication topology that arises from maximizing the mutual information between the optimal actions and the combined available measured variables. 
For case 1, where we minimize voltage deviations and losses across the network with DERs spread randomly across all buses, we see no particular patterns in the optimal communication infrastructure.
Some DERs at the end of the network tend to communicate with loads nearby, which can be explained by supply tending to match demand nearby to minimize losses higher upstream in the radial network.
For case 2, where we are localizing the production of power by minimizing the power delivered from the substation, an interesting pattern arises from the fact that the loads and DER are concentrated in \emph{different} areas.
As a result, if we allow DERs to observe one extra node anywhere in the network, all select a node in the \emph{other} area with concentrated loads. This makes intuitive sense as the optimal control actions are largely responding to the high demand in the other area.  
In addition, across all 20 DERs the selected nodes are all in a set of only 3 nodes, suggesting an efficient sensor implementation.
A relevant question that arises when communication is needed, is whether the different agents have the economic incentives to signal their information truthfully. Here, we have assumed that the inverters are all managed by one entity. However, in the practical scenario where third-parties operate different controllers, a further analysis of how signaling incentives may affect the quality of the decentralized control scheme is relevant, and inspiration may be drawn from the literature on optimal decentralized control~\cite{mahajan_information_2012}.

\subsection{Interpreting the Machine Learning Models}
\label{subsec:interpretation}
In this section, we analyze how the structure of different learned controllers~$i,j \in \C$ differs across the network as reflected by parameters~$\beta^{(i)},\beta^{(j)}$ in~\eqref{eq:MLmodel}. What does this tell us about the network and OPF problem?

In our first case study, we determined regression models for 27 different inverters. Table \ref{tab:InvMdl} shows regression results for inverters A and B (both depicted in Figure~\ref{fig:FeederDiag}). The first two columns present the  features selected by the stepwise regression and the values for the $\beta$-coefficients in~\eqref{eq:MLmodel}. The third column shows the standard error of the estimate, and the fourth lists p-values, which here means the probability that a coefficient is zero. 
Note how the stepwise regression approach results in two clearly different models.
The reactive power output of inverter A depends predominantly on the local reactive power consumption $\varphi^{(A)}_2$, while the output of inverter B is strongly related to the reactive power capacity $\varphi^{(B)}_3$ and $\varphi^{(B)}_2$ is less relevant.
Inverter A's structure can be explained by the fact it is located at the end of the feeder. As a result, the aggregate power flow at A is relatively low and correlated with A's own consumption. As the objective is to minimize losses and voltage deviations, A tries to produce a signal that looks like providing the neighboring demand for reactive power, thereby minimizing current locally \emph{and} upstream.
Inverter B's structure can be explained by its location close to the feeder head. In this area, the aggregate power flow is much larger, and the output of inverter B is needed more as a ``bulk'' product that lowers the loss and voltage drop on the branch between B and the substation.
This causes inverter B to operate at its maximum capacity in OPF most of the time, causing correlation of dynamic capacity and the optimal setpoint.
This example illustrates that optimal reactive power output of inverters tend to have idiosyncratic structures depending on location and the assignment of loads, generation and capacity across a network. Therefore, effective design of controllers based on local measurements is complex, and rather than relying on one-size-fits-all solutions, better and safer results may be achieved with a flexible data-driven approach that respects constraints. 
%
\begin{table}[!t]
\renewcommand{\arraystretch}{1.3}
\caption{Normalized regression coefficients for inverter A and B, as depicted in Figure~\ref{fig:FeederDiag}.}
\label{tab:InvMdl}
\centering
\resizebox{\columnwidth}{!}{%
\begin{tabular}{c c c c || c c c c}
\hline
A & Est. & SE & p-value &  B & Est. & SE & p-value \\
\hline\hline
offset	&	0.02	&	0.01	&	0.01				&	offset	&	0.02	&	0.00	&	$8.2 e^{-4}$ \\
$\varphi_1$	&	0.37	&	0.01	&	$0$		&	$\varphi_1$	&	0.04	&	0.00	&	$0$ \\
$\varphi_2$	&	0.77	&	0.01	&	0				&	$\varphi_3$	&	0.96	&	0.01	&	0 \\
$\varphi_3$	&	-0.21	&	0.02	&	$0$		&	$\varphi_1 \varphi_3$	&	0.06	&	0.01	&	$0 $\\
$\varphi_1 \varphi_2$	&	0.17	&	0.01	&	$0$		&	$\varphi_1^2$	&	-0.03	&	0.00	&	$0$\\
$\varphi_1^2$	&	-0.06	&	0.01	&	$0$		&	$\varphi_3^2$	&	-0.03	&	0.01	&	$0$ \\
\hline
\end{tabular}%
}
\end{table}


\subsection{Designing Across Planning and Operation}
\label{subsec:design}
When using supervised learning, some immediate concerns arise around the quality of the training set and the historical data used to construct it.
Firstly, a designer wants to have certainty that the historical data reflect the ``normal'' system behavior expected in future operation. If certain scenarios are not measured historically, the controllers may not \emph{learn} how to respond in an optimal or desired way. 
If future scenarios are not measured, but can be anticipated, it may be possible to instead \emph{simulate} these and augment these to the historical data.
Secondly, a designer will want to analyze how the system behaves under faults and safety-critical scenarios or want to know what happens when changes occur in the system, such as a new street block connecting to a feeder or smaller increments, such as the installation of new electric vehicles. Based on the effects on the behavior of the controllers, it may or may not be needed to retrain all local policies through the central OPF and machine learning steps as covered in Sections~\ref{sec:opf} and~\ref{sec:regression}, and send new controller parameters to the inverters. 
In our future work, we are assessing different learning-based controllers for OPF on their performance across a spectrum of system changes~\cite{karagiannopoulos_data-driven_2019}. For smaller changes, a periodic retraining procedure can update the control parameters incrementally by augmenting the training data with more recent historical measurements.

\section{Conclusions}
\label{sec:conclusions}

In this paper, we presented an integrative approach for decentralizing optimal power flow (OPF) methods based on machine learning. 
By exploiting the information in local exogenous disturbance variables, a machine learning model is able to reconstruct the optimal set point to OPF problems. In concert, all machine learning models implemented in a decentralized fashion mimic the execution of a centralized OPF problem.
A rate distortion framework allowed to interpret the decentralized learning approach as a compression and reconstruction problem, providing a theoretical lower bound on the distortion that can be achieved with local information and a procedure to improve controllers by communicating with a node that maximizes the mutual information between the optimal control and the resulting available variables. 
Experiments on both single- and three-phase unbalanced (see supplementary material) systems illustrated the relevancy of Decentralized OPF for a broad range of systems and objectives.
Lastly, we provided analysis and discussion about the effect of communication, the interpretation of controller parameters and the process of updating the controllers in an ongoing fashion.






\section*{Acknowledgments}
We thank Michael Sankur for contributions to the code base for this work. This work is supported by the National Science Foundation’s CPS FORCES Grant (award number CNS-1239166) and the UC-Philippine-California Advanced Research Institute (award number IIID-2015-10).


\ifCLASSOPTIONcaptionsoff
  \newpage
\fi



%
 \bibliographystyle{IEEEtran}
 \bibliography{references_def.bib} 
\end{document}